\documentclass{article}

\usepackage{arxiv}

\usepackage[utf8]{inputenc} 
\usepackage[T1]{fontenc}    
\usepackage{hyperref}       
\usepackage{url}            
\usepackage{booktabs}       

\usepackage{mathtools}
\usepackage{amsthm,amssymb}
\usepackage{nicefrac}       
\usepackage{microtype}      
\usepackage{cleveref}       
\usepackage{lipsum}         
\usepackage{graphicx}

\usepackage{doi}

\usepackage{xcolor}
\usepackage{xspace}

\DeclareMathOperator*{\argmax}{argmax}
\DeclareMathOperator*{\argmin}{argmin}

\renewcommand\[{\begin{equation}}
\renewcommand\]{\end{equation}}

\newcommand{\Ind}[1]{\ensuremath{\mathbbm{1}\!\left\{#1\right\}}}

\newcommand{\Reg}{\ensuremath{\mathrm{Reg}\xspace}}

\newcommand{\Loss}{\ensuremath{\mathrm{Loss}\xspace}}

\theoremstyle{plain}

\include{definitions}

\title{Machine Learning for Combinatorial Optimisation of Partially-Specified Problems: Regret Minimisation as a Unifying Lens}

\date{}

\author{
    Stefano Teso \\
	University of Trento \\
	\texttt{stefano.teso@unitn.it} \\
	\And
    Laurens Bliek \\
    TU Eindhoven \\
    \texttt{l.bliek@tue.nl} \\
    \And
    Andrea Borghesi \\
    University of Bologna \\
    \texttt{andrea.borghesi3@unibo.it} \\
    \AND
    Michele Lombardi \\
    University of Bologna \\
    \texttt{michele.lombardi2@unibo.it} \\
    \And
    Neil Yorke-Smith \\
    TU Delft \\
    \texttt{n.yorke-smith@tudelft.nl} \\
    \And
    Tias Guns \\
    KU Leuven \\
    \texttt{tias.guns@kuleuven.be} \\
    \AND
    Andrea Passerini \\
    University of Trento \\
    \texttt{andrea.passerini@unitn.it}
}

\renewcommand{\headeright}{}
\renewcommand{\undertitle}{}
\renewcommand{\shorttitle}{Machine Learning for Combinatorial Optimisation of Partially-Specified Problems}

\hypersetup{
    pdftitle={Machine Learning for Combinatorial Optimisation of Partially-Specified Problems},
    pdfauthor={Stefano Teso, Laurens Bliek, Andrea Borghesi, Michele Lombardi, Neil Yorke-Smith, Tias Guns, Andrea Passerini},
    pdfkeywords={Machine Learning, Combinatorial Optimisation, Regret Minimisation, Surrogate-based Optimisation, Empirical Model Learning, Decision-focused Learning, Structured-output Prediction},
    colorlinks=true,
    citecolor=blue,
    linkcolor=black,
    filecolor=black,
    urlcolor=blue,
}

\begin{document}
\maketitle

\begin{abstract}
    It is increasingly common to solve combinatorial optimisation problems that are partially-specified. 
    We survey the case where the \textit{objective function} or the \textit{relations between variables} are not known or are only partially specified.
    The challenge is
    to learn them from
    available data, while taking into account a set of hard
    constraints that a solution must satisfy, and that solving the
    optimisation problem (esp.\@ during learning) is computationally
    very demanding.  This paper overviews four seemingly unrelated
    approaches, that can each be viewed as learning the objective
    function of a hard combinatorial optimisation problem: 1)
    surrogate-based optimisation, 2) empirical model learning, 3)
    decision-focused learning (`predict + optimise'), and 4)
    structured-output prediction.  We formalise each learning
    paradigm, at first in the ways commonly found in the literature,
    and then bring the formalisations together in a compatible way using regret.
    We
    discuss the differences and interactions between these frameworks,
    highlight the opportunities for cross-fertilization and
    survey open directions.
\end{abstract}


\section{Introduction}
\label{sec:intro}

In the burgeoning research area of machine learning (ML) for combinatorial optimisation (CO), attention is now coming to \emph{partially-specified problems}: those for which the problem instance itself is not fully known.
This survey focuses on such problems where we must learn the objective
function and/or the relationships between variables, while we assume
the hard constraints to be given (see~\cite{de2018learning} for a survey on
constraint learning).
Such instances are commonplace in, for example, vehicle routing (driver preferences), multi-objective optimisation (weight penalties), and energy-aware scheduling (energy prices).

We survey four seemingly unrelated
lines of work, that can each be viewed as learning the objective
function of a hard combinatorial optimisation problem: 1)
surrogate-based optimisation, which faces expensive-to-evaluate
objective functions.  2) empirical model learning, which embeds ML
models in the CO problem to empirically deal with hard-to-model,
complex components; 3) decision-focused learning (`predict +
optimise'), which learns the cost coefficients of CO problem in a
decision-focused way; and 4) structured-output prediction, which
learns the objective function from a dataset of input-output pairs.

Existing surveys look at ML for (augmenting) CO solving \cite{DBLP:journals/eor/BengioLP21}; at ML for (augmenting) CO modelling \cite{DBLP:conf/ijcai/0001M18}; at end-to-end CO learning, in particular predict-and-optimise \cite{DBLP:conf/ijcai/KotaryFHW21}; and at graph neural networks for CO \cite{DBLP:conf/ijcai/CappartCK00V21}.  What is missing is a systematic survey that compares the interaction between ML-for-CO frameworks on a common basis, particularly for partially-specified problems. This paper fills that gap in the literature by formalising the above four lines of work on a common basis, namely \emph{regret minimisation}.  This unified view facilitates knowledge transfer between the problems (as we highlight in Section~\ref{sec:interactions}) and encourages concerted research.

\section{Four Problems Types in ML-for-CO}
\label{sec:problems}

\begin{table}[tb]
    \begin{small}
    \begin{center}
    \begin{tabular}{ll}
        \toprule
        \textbf{Symbol} & \textbf{Meaning} \\
        \midrule
        $f$             & Objective function of the COP \\
        $\calC$         & Hard constraints of the COP \\
        $h^*$           & Complex function appearing in the COP \\
        $\hat{h}$       & Surrogate of the complex function \\
        $\theta^*$      & Ground-truth parameters of $f$ \\
        $\hat{\theta}$  & Estimated parameters \\
        \midrule
        $x$             & Input to the COP, often observed features \\
        $\calY$         & Search space of the COP defined by $\calC$ \\
        $y^*$           & Optimal COP solution using $\theta^*$ and/or $h^*$ \\
        $\hat{y}$       & Optimal COP solution using $\hat{\theta}$ and/or $\hat{h}$ \\
        \midrule
        $\Reg^*$        & Regret measured w.r.t.\@ $f$ using $\theta^*$ and/or $h^*$  \\
        $\widehat{\Reg}$& Regret measured w.r.t.\@ $f$ using $\hat\theta$ and/or $\hat{h}$ \\
        \bottomrule
    \end{tabular}
    \end{center}
    \end{small}
    \caption{Notation used throughout the paper.}
    \label{tab:notation}
\end{table}

We consider four learning and optimisation tasks revolving around generic combinatorial optimisation problems (COPs).  In short, a COP aims to minimise an objective function $f(y)$ subject to hard constraints $\calC$:
\begin{equation}
    \argmin_{y \in \calY} f(y).
    \label{eq:COP}
\end{equation}
Here, $y$ are discrete or continuous-discrete decision variables and $\calY$
the (implicit) set of all feasible assignments to $y$ consistent with $\calC$.
Table~\ref{tab:notation} summarises the notation used.

\subsection{Surrogate-Based Optimisation (SBO)}

SBO algorithms such as Bayesian optimisation~\cite{movckus1975bayesian,Shahriari2016TakingTH} are concerned with
\begin{align}
    \argmin_{y \in \mathcal Y} \ h^*(y),
    \label{eq:SBO}
\end{align}
where the objective function $h^*$ is assumed to be expensive, non-deterministic, non-convex, nonlinear, black-box, etc. 
That is, the complex function in this optimisation problem is the objective function itself.
This problem setting appears in, for example, automatic algorithm configuration and computational fluid dynamics.
The problem~\eqref{eq:SBO} is typically solved in an \emph{iterative} manner, choosing carefully which candidate solution $y$ to try in each iteration,
to evaluate the expensive objective $h^*$ less often.

At every iteration $n$ of a SBO algorithm, we assume 
we have 
evaluated $h^*$ for $n-1$ candidate solutions, i.e., we have access to a dataset $D_{n-1} = \{(y_i, h^*(y_i)\}_{i=1}^{n-1}$.
Instead of optimising $h^*$ directly, a \emph{surrogate model} $\hat h$ is learned from the data by minimising a loss function in a supervised learning approach:
\begin{equation}
    \hat h = \argmin_h \ \mathbb{E}\left[ \Loss(h,D_{n-1}) \right].     \label{eq:minloss_sbo}
\end{equation}
The loss function can consist of the 
MSE, though typically a Bayesian framework is used.
During training, the choice of which candidate solution to evaluate (and hence learn from) next is made by maximising an acquisition function $\alpha$:
\begin{align}
    \argmax_{y \in \mathcal{\hat Y}} \ \alpha(\hat h(y)).
\end{align}
The acquisition function
is used to balance the trade-off between exploration and exploitation.
Example acquisition functions are
Expected Improvement, Upper Confidence Bound, Thompson sampling~\cite{Shahriari2016TakingTH}, and Entropy Search~\cite{Wang2017MaxvalueES}.
As some of the constraints could be incorporated in the acquisition function or otherwise relaxed, the new search space $\mathcal{\hat Y}$ could differ from the original.

SBO is typically used in continuous optimisation with simple bound constraints, e.g., $\mathcal Y = [0,1]^d$, but combinatorial SBO has recently gathered attention.
Examples of recent SBO approaches that deal with diskernels~\cite{zaefferer2018Auth,oh2019combinatorial,Deshwal2021BayesianOO},
kernels~\cite{zaefferer2018Auth,oh2019combinatorial},
rounding~\cite{luong2019bayesian}, semi-definite programming and sub-modular relaxations~\cite{baptista2018bayesian,deshwal2020scalable}, piece-wise linear or polynomial models~\cite{Daxberger2020MixedVariableBO,MVRSM}, or
bandits~\cite{ru2019bayesian}.

\subsection{Empirical Model Learning (EML)}

EML
refers to a family of techniques that aim at `embedding' ML models in declarative optimisation.  The term \emph{embedding} refers to any technique that enables the declarative solver to reason on the ML model, together with the rest of the problem structure.
The approach is chiefly motivated by decision-making with complex systems, which are not well suited for a traditional, declarative modelling approach. Applications include cost minimization in energy policies under adoption constraints, thermal-aware job allocation, or planning non-pharmaceutical interventions for epidemic control.  The focus is on problems in the form \cite{DBLP:conf/ijcai/0001M18}:
\begin{equation}
    \argmin_{y \in \mathcal{Y}, y_{out} = h^*(y_{in})} f(y)
\label{eq:eml:truth}
\end{equation}
where $h^*$ represent the complex, hard-to-model, system, and $\calY$ defines a set of constraints. The decision variables $y$ are partitioned in two groups, i.e. $y = (y_{in}, y_{out})$, where $y_{in}$ refers to those representing the input of the complex system $h$, and $y_{out}$ to the output. Variables that are not directly tied to $h$ can be considered part of $y_{in}$ to simplify the notation.

As per the stated assumptions, $\mathcal{Y}$ can be modelled via traditional constraints, while that is not possible for $h^*$.
EML suggests to deal with this issue by replacing $h^*$ with a surrogate model $\hat{h}$ obtained via ML, thus yielding:
\begin{equation}
\mathop{\argmin}_{y \in \mathcal{Y},y_{out} = \hat{h}(y_{in})} \ f(y)
\label{eq:eml:viable}
\end{equation}
where training is assumed to be performed \emph{prior} to optimisation (unlike in SBO).
The focus
is then
making the surrogate model $\hat{h}$ manageable for the chosen declarative solver.

The adopted technique depends on the type of optimisation technology.  In mathematical programming (e.g., MIP, MINLP), it is typical to encode the ML model by 
standard modelling constructs offered by the formalism: this approach was employed for ReLU neural networks 
in \cite{DBLP:journals/constraints/FischettiJ18}, for feed-forward NNs with sigmoid activation in \cite{DBLP:journals/ai/LombardiMB17}, and more recently for NNs with piecewise activation functions in \cite{anderson2020strong}.  In constraint programming, it is possible to use global constraints to model the entire ML component, by providing ad hoc filtering algorithms: this is the approach taken in \cite{DBLP:journals/constraints/LombardiG16,DBLP:journals/ai/LombardiMB17} for neural networks and in \cite{DBLP:conf/cpaior/BonfiettiLM15} for decision trees.  In some cases, changes to the solution engine itself are proposed in an effort to improve performance: examples include the branching heuristic from \cite{DBLP:conf/cav/KatzBDJK17} and the cut generation process from \cite{anderson2020strong}.
While the EML approach was initially motivated by the need to deal with hard-to-model systems, the method can be (and has been) employed for the verification of ML models and the generation of counter-examples, e.g., in \cite{DBLP:journals/constraints/FischettiJ18}.

\subsection{Decision-Focused Learning (DFL)}

DFL considers a COP in which the structure of the objective function $f$ is fixed, e.g., it is a linear or quadratic function, but the coefficients $\theta$ of $f$ are not known. For example, for a linear function $f(y;\,\theta) = \theta^\intercal y$ with $|\theta| = |y|$. The COP hence takes the following form,
where the values of the ground-truth parameters $\theta^*$ are unknown but we assume access to correlated features $x$ and a historic dataset $D = \{(x_i, \theta^*_i)\}_{i=1}^n$:
\begin{equation}
    \argmin_{y \in \calY} f(y;\,\theta^*)
    \label{eq:DFL:COP}
\end{equation}
From this, a straightforward approach is to learn a function that predicts $\theta^*$ from this data, i.e., treat it as a multi-output regression problem.
This model can be learned by fitting the data $D$ to minimise some loss function:
\begin{equation}
\textstyle
\argmin_\omega \mathbb{E} \left[ \Loss(\hat{\theta}(x),\theta^*) \right] \label{eq:minregret}
\end{equation}

\noindent where $\omega$ are the learned parameters of the predictor $\hat{\theta}(x)$; an example loss is the Mean Squared Error. The predicted cost vectors can then be used to solve Eq~\eqref{eq:DFL:COP} in a second phase.

However, it has been shown that better results can be obtained by learning a function for $\theta$ that directly takes into account its effect on the optimisation problem \cite{elmachtoub2021smart,aaai/WilderDT19,mandi2020interior}.

In order to measure the effect of a predictive model of $\theta$ on the COP, we compute the \emph{regret}: the difference in actual cost between the optimal solution for the estimated parameter values $\hat{y}$ when optimising with $\hat{\theta}$, and the true value of the optimal solution $y^*$ for the true parameter values $\theta^*$:
\begin{equation}
\Reg(\hat{\theta},\theta^*) = f(\hat{y};\,\theta^*) - f(y^*;\,\theta^*) 
\end{equation}
where obtaining $\hat{y}$ requires solving 
Eq.~\eqref{eq:DFL:COP} with $\hat{\theta}$.

The goal of decision-focussed learning is then to learn a model $\hat\theta$ that minimises the regret of the resulting predictions
\begin{equation}
\argmin_\omega \ \mathbb{E}\left[ \Reg(\hat{\theta}(x),\theta^*) \right]. \label{eq:dfl-as-regret-minimisation}
\end{equation}
However, regret cannot be directly used as a loss function because it is non-continuous and involves differentiating over the argmin in Eq.~\eqref{eq:DFL:COP}.
Instead, the problem can be cast as a bi-level optimisation, with the master problem minimising over $\omega$ and the subproblems 
optimising $\hat{y}$.
More effective are gradient-descent based methods that define `implicit differentiation' techniques for quadratic and Mixed Integer Programming 
\cite{aaai/WilderDT19,ferber2020mipaal,mandi2020interior}, or that use sub-gradient methods based on perturbing the predictions and comparing the true solution and solution with perturbed predictions~\cite{elmachtoub2021smart,PogancicPMMR20,mulamba2020discrete}.

\subsection{Structured-Output Prediction (SOP)}
\label{sec:sop}

The goal of SOP is to predict an output structure $y \in \calY$ from an input $x \in \calX$, also typically structured~\cite{deshwal2019learning}.
Prototypical applications
include part-of-speech tagging,
protein secondary structure prediction,
and floor layout synthesis.

At a high level, SOP requires to learn a mapping $F: \calX \to \calY$ from a dataset of ground-truth pairs $\{ (x_i, y_i^*) \}_{i=1}^m$ where $x_i \in \calX$ are observed inputs and $y_i^* \in \calY$ are associated optimal outputs.

Since learning $F$ directly is non-trivial, \emph{energy-based} predictors~\cite{LeCun06atutorial} instead acquire an energy function $f : \calX \times \calY \to \bbR$, with parameters $\theta$, that measures the compatibility between input and output.  Predicting an output for a given $x$ amounts to finding one with lowest energy:
\begin{equation}
    F(x;\theta) = \argmin_{y \in \calY} f(x,y;\theta).
    \label{eq:structured_prediction}
\end{equation}
In this paper, we consider the challenging setting where $\calY$ is defined by hard feasibility constraints $\calC$ encoding, e.g., chemical validity rules or engineering requirements. In this case, Eq.~\ref{eq:structured_prediction} can be naturally viewed as a COP.
Then, learning aims at finding parameters $\hat\theta$ that associate minimal energy to the correct output $y_i^*$ of each training input $x_i$.

Traditionally, $f$ has been implemented as a linear function $f(x,y;\theta) = \theta^\intercal \Psi(x,y)$, where $\Psi$ denotes joint features of the input-output pair;
inference (Eq.~\ref{eq:structured_prediction}) is solved with dynamic programming~\cite{joachims2009predicting}, or ILP~\cite{pan2018} and SMT~\cite{lion11,teso2017structured} solvers;
learning can be tackled using cutting planes~\cite{joachims2009predicting}, Frank-Wolfe~\cite{lacoste2013block}, and sub-gradient descent~\cite{ratliff2007approximate}.
Intuitively, these strategies refine the current estimate $\hat\theta$ by iteratively identifying the most offending incorrect answer:
\[
    \hat{y}_i = \argmin_{y\in\calY, \ y \ne y_i^*} f(x_i, y; \hat\theta)
    \label{eq:sop-moia}
\]
and increasing its associated energy.
A closely related setup, common in human-in-the-loop applications, is to learn -- either directly~\cite{dragone2018constructive} or indirectly~\cite{shivaswamy2015coactive} -- from ranking feedback of the form $(x_i, (y_{i,\mathrm{better}}^* \succeq y_{i,\mathrm{worse}}^*))$. 
Many approaches for SOP with deep energy functions focus on
continuous relaxations of the original
problem~\cite{chen2015,Zheng2015}, that are however inherently
sub-optimal for CO~\cite{Raghavendra2008}.
A notable exception is gradient-based inference~\cite{Lee2019}, which
encourages unconstrained inference to produce feasible solutions by adjusting the network
weights at test time.
A promising 
line of work incorporates
COP solvers as differentiable layers in neural
architectures~\cite{ferber2020mipaal}, where differentiability is achieved via 
continuous interpolations~\cite{rolinek2020,Vlastelica2020} or perturbation-based
implicit differentiation~\cite{niepert21imle}.

\begin{table*}[tb]
    \centering
    \begin{small}
    \begin{tabular}{clll}
        \toprule
        \textbf{Problem}
            & \textbf{COP}
            & \textbf{Supervision}
            & \textbf{Learn}
        \\
        \midrule
        SBO
            & $\argmin_{y \in \calY} h^*(y)$
            & $\{ (y_i,\, h^*(y_i)) \}$
            & $\hat{h}(\cdot)$ that gives low-regret outputs
        \\
        EML
            & $\argmin_{y \in \mathcal{Y}} f(y) \ \mathrm{s.t.} \ y_{out} = h^*(y_{in})$
            & $\{ (y_{i,in}, \, h^*(y_{i,in})) \}$
            & $\hat{h}(\cdot)$ that gives low-regret outputs
        \\
        DFL 
            & $\argmin_{y \in \calY} f(y;\, \theta^*)$
            & $\{(x_i,\, \theta^*(x_i)) \}$
            & $\hat{\theta}(x_i)$ that gives low-regret outputs
        \\
        SOP
            & $\argmin_{y \in \calY} f(x, y;\, \theta^*)$
            & $\{ (x_i,\, y_i^*) \}$
            & $\hat{\theta}$ that gives low-regret outputs
        \\
        \bottomrule
    \end{tabular}
    \end{small}
    \vspace{-0.5\baselineskip}
    \caption{Summary of the ML-for-CO problems we consider.}
    \label{tab:comparison}
\end{table*}

\section{A Unified View via Regret Minimisation}
\label{sec:unification}

\noindent
We now have everything together to bridge the different problems surveyed in the previous section through the lens of \emph{regret minimisation}.
We start by providing a unified formulation of their underlying COPs and showing that it naturally leads to a \emph{ground-truth} regret minimisation problem.
Then we proceed to cast common solution strategies in terms of (\emph{ground-truth} or \emph{model-side}) regret minimisation, making it possible to see them as (approximate) strategies for solving the newly introduced unified formulation.
Many interesting links unveiled by this perspective are discussed in the next section.

\subsection{A Unified Inference Problem}

Table~\ref{tab:comparison} summarises the various COP problems and learning problems
using a common notation. Written in this format, it becomes clear that we can generalise their respective optimisation problems into the following \emph{unified COP}:
\begin{equation}
    \tag{UCOP}
    y^* = \argmin_{y \in \calY,y_{\mathit{out}} = h^*(y_{\mathit{in}})} \ f(x, y; \, \theta^*(x)).
    \label{eq:ucop}
\end{equation}
Indeed, EML can be retrieved from the \ref{eq:ucop} by assuming both $\theta^*$ and $x$ to be known (e.g. since we are dealing with fixed problem instances). SBO can be retrieved by removing $y_{\mathit{out}}$, and further assuming $f(x, y; \theta^*(x)) = h^*(y)$, i.e. $f$ is the complex component.
Similarly, UCOP reduces to DFL when no complex component $h^*$ is present. SOP can be derived by further assuming that is $f$ reasonably efficient to optimize and that $\theta^*$ does not depend on $x$.

From this perspective, we see that all four problems are concerned with acquiring a constrained optimisation problem, obtained by replacing $\theta^*$ and/or $h^*$ with their respective estimates $\hat\theta$ and $\hat{h}$ in the \ref{eq:ucop}, that can be optimised efficiently and whose solutions have ground-truth cost comparable to that of the \ref{eq:ucop}. Formally, this equates to requiring the solutions of the approximate COP to attain low \emph{ground-truth regret}, defined for arbitrary $y$ and $y^{\prime}$ as
\[
    \Reg^*(x,y,y^\prime) = f(x,y^\prime;\,\theta^*) - f(x,y;\,\theta^*).
    \label{eq:ground-truth-regret}
\]
The ground-truth regret cannot be minimised directly 
when
$\theta^*$ is not known (one exception being DFL), or due to the presence of the complex component $h^*$.  An alternative measure of regret that works around this restriction is introduced below.

\subsection{Unifying the Solution Strategies}

Any solution strategy must directly or indirectly optimise the quality of the solutions output by \ref{eq:ucop}.  It turns out that existing strategies for all four problems can be cast in these terms: DFL is already explicitly written in this form (see Eq.~\ref{eq:dfl-as-regret-minimisation}), while strategies for the other problems can be converted.

\subsubsection{SOP as Regret Minimisation}
\label{sec:sop-as-regret-minimisation}

The role of regret in energy-based approaches to SOP is not explicit, but it can be uncovered as follows.
Let $\hat{y}_i$ be the most offending incorrect answer for an example $(x_i, y_i^*)$, as per Eq.~\ref{eq:sop-moia}.
Then, the overall SOP learning problem becomes:
\[
    \min_{\hat\theta} \frac{1}{m} \sum_i \big| f(x_i, y_i^*; \hat\theta) - f(x_i, \hat{y}_i; \hat\theta) \big|_+
    \label{eq:sop-as-sum}
\]
where $|s|_+ = \max\{0, s\}$. Now, if the estimated energy of $y_i^*$ is \emph{lower} than that of $\hat{y}_i$, then the example is classified correctly and it does not contribute to Eq.~\ref{eq:sop-as-sum}.  Otherwise, the model wrongly predicts $\hat{y}_i$, i.e., $\hat{y}_i \in F(x_i; \hat\theta)$.  Hence, we can rewrite Eq.~\ref{eq:sop-as-sum} into the following regret minimisation problem:
\[
    \argmin_{\hat\theta} \ \frac{1}{m} \sum_i \ \widehat{\Reg}(x_i, y_i^*, F(x_i; \hat\theta)).
    \label{eq:sop-as-regret-minimisation}
\]
Note $F$ might be non-unique.
Here, $\widehat{\Reg}$ is the \emph{model-side regret} and it measures the difference between the cost that the \emph{model} assigns to to the ground-truth $y_i^*$ and the cost that it assigns to its own predictions $\hat{y}_i \in F(x_i; \hat\theta)$:
\[
    \textstyle
    \widehat{\Reg}(x, y, y^\prime) = f(x, y; \hat\theta) - f(x, y^\prime; \hat\theta).
    \label{eq:sop-model-regret}
\]
As long as $\hat{y}$ is a minimiser of $f$ (i.e., Eq.~\ref{eq:sop-moia} is solved to global optimality), then the model-side regret is non-negative.
If not (e.g., because $\hat{y}$ is computed using approximate inference), some learning algorithms may face difficulties~\cite{shivaswamy2015coactive}.

Essentially, the model-side regret works as a surrogate for the real objective of the learning problem, namely the ground-truth regret.
Naturally, in SOP the parameters $\theta^*$ are unobserved, and as such the ground-truth regret cannot be computed nor optimised.  This is a crucial difference with DFL, which instead has access to the ground-truth regret, cf. Eq.~\ref{eq:dfl-as-regret-minimisation}.

Eq.~\ref{eq:sop-as-regret-minimisation} encourages the model to prevent incorrect outputs $y^\prime \ne y_i$ from costing less than the optimal 
$y_i$.  A widely-used variant requires that all wrong 
incorrect answers have a \emph{larger enough} cost.  This is done by replacing Eq.~\ref{eq:sop-moia} with:
\[
    \argmax_{y\in\calY, \ y \ne y_i^*} \ f(x_i, y_i^*; \theta) - f(x_i, y; \theta) + \Delta(y_i^*, y).
    \label{eq:sop-moia-with-margin}
\]
and $| f(x_i, y_i^*; \hat\theta) - f(x_i, \hat{y}_i; \hat\theta)|_+$ with  $|f(x_i, y_i^*; \hat\theta) - f(x_i, \hat{y}_i; \hat\theta) + \Delta(y_i^*, \hat{y}_i)|_+$ in Eq.~\ref{eq:sop-as-sum}. Here $\Delta(y^*, y)$ is a user-supplied function that measures how much $y$ differs from $y^*$~\cite{joachims2009predicting}.
In the simplest case, it can be used to encourage wrong configurations to have a cost difference of at least one, that is, $\Delta(y, y_i) = \Ind{y \ne y_i}$.  More refined alternatives take into consideration the functional or perceptually dissimilarity between outputs.
Thanks to $\Delta$, the learned model is encouraged to separate optimal and sub-optimal outputs by a larger margin, which tends to improve generalisation to unseen instances~\cite{joachims2009predicting}.

\subsubsection{EML as Regret Minimisation}
\label{sec:eml-as-regret-minimisation}

EML is usually presented as learning a black-box model from appropriate supervision and then plugging the latter into an optimisation problem, but it is possible (and appropriate) to frame it in terms of regret.
Let $y^*$ refer to the solution of the (computationally infeasible) ground-truth problem in EML. Then the goal of EML can be rephrased as minimising the gap between the evaluations of the two solutions w.r.t.\@ the true system behaviour. Formally, equation~\eqref{eq:eml:viable} can be rewritten as:
\[
     \hat{y} = \argmin_{y \in \calY, y_{\mathit{out}} = \hat{h}(y_{\mathit{in}})} \ \Reg^*(x,y,y^*).
\]
The equivalence holds since $y^*$ is fixed, making $y$ the only practically relevant variable. In principle, infinite regret is possible if hard constraints are violated, but this is usually symptomatic of a poor problem formulation: if feasibility is an issue, it may be more appropriate to use soft constraints, which can be considered part of the cost function.

The new notation highlights that, even if the surrogate model $\hat{h}$ makes mistakes, those do not matter as long as the resulting solution is close enough to the true optimum.  This behaviour has been noticed before \cite{DBLP:journals/ai/LombardiMB17}, but never formally characterised.

\subsubsection{SBO as Regret Minimisation}
\label{sec:sbo-as-regret-minimisation}

SBO solves problems in the form of $y^* = \argmin_{y\in\mathcal Y} h^*(y) $ by swapping the hard-to-optimise $h^*$ function with a more tractable surrogate $\hat h$, which is learned by minimising the loss as in~\eqref{eq:minloss_sbo}. Accordingly, the original problem is changed to $ \hat y = \argmax_{y\in\mathcal{\hat Y}} \alpha(\hat h(y))$, for some acquisition function $\alpha$. 
Instead of minimising loss when learning the surrogate, we can formulate the learning problem as minimising the regret:
\begin{align}
    \textstyle
    \argmin_{\hat h} \ \Reg^*(\hat h) & = 
    h^*(\hat y) - h^*(y^*).
    \label{eq:reg_sbo}
\end{align}
The regret, which is similar to what is called \emph{inference regret} in the literature~\cite{Wang2017MaxvalueES}, is non-negative under the assumption that $\hat y \in \mathcal Y$.
Note that other measures of regret exist, and these other forms (or their approximations) are often incorporated in the acquisition function $\alpha$~\cite{Shahriari2016TakingTH} or in regret bounds~\cite{srinivas2012information}.
Especially when the global optimum $h^*(y^*)$ is known, even if $y^*$ is unknown, it is possible to incorporate these other regrets 
in the acquisition function $\alpha$ directly \cite{Nguyen2020KnowingTW}.
However, to the best of our knowledge, especially when looking at combinatorial problems, there exists no SBO approach that takes the regret in~\eqref{eq:reg_sbo} into account when \emph{learning} the surrogate $\hat h$.
This is for good reason: even under the assumption that $h^*(y^*)$ is known, $\hat y$, and also $h^*(\hat y)$, are not easy to compute, despite $\hat y$ depending on the easy-to-evaluate surrogate model rather than the expensive objective.
An interesting direction for solving this new learning problem could be considering surrogate models (and acquisition functions) for which $\hat y$ 
is easy to compute, such as models that are convex in $y$, though such a surrogate might suffer from a loss of accuracy.
Exploratory work on such input-convex surrogate models can be found in e.g.~\cite{CDONEpaper}.

\section{Discussion, Synergies and Outlook}
\label{sec:interactions}

\paragraph{SOP vs DFL}
A first outcome of our unified formulation is the observation that SOP and DFL are quite close, as they both assume all elements of the \ref{eq:ucop} to be known \emph{except} for the ground-truth parameters $\theta^*$, and that $f$ can be repeatedly (approximately) optimised during training at a reasonable computational cost.  This means that -- in stark contrast to the other two problems -- in SOP and DFL it is not necessary to acquire and use a surrogate of the COP.
The major differences between the two problems are whether the ground-truth parameters $\theta^*$ are independent from the input $x$ (SOP) or not (DFL), and in what supervision is available.  Indeed, in SOP the predictor has no access to $\theta^*$, while in DFL this information is available for all training examples.
This turns out to be a key difference, as also noted by~\cite{elmachtoub2021smart}, because it entails that algorithms for SOP are restricted to computing the model-side regret, while those for DFL can evaluate and optimise also the ground-truth regret.  Naturally, optimising the latter more directly improves the 
down-stream solutions.

\paragraph{SBO vs EML}
EML and SBO are also tightly related:  they both assume the whole optimisation problem to be available from the start.  Moreover, the only supervision they require provides information about the output of the complex component $h^*$.
This supervision is built iteratively in SBO while learning the surrogate $\hat h \approx h^*$: the method carefully chooses which candidate to try next based on the information obtained in previous iterations, and collects additional ground truth labels in the process. Conversely, EML has historically relied on pre-collected datasets.

A major difference with DFL and SOP is that, since $h^*$ is hard to optimise, the optimum $y^*$ of the optimisation problem cannot be easily recovered. In fact, finding the optimal output of the complex component $h^*$ is what constitutes the entire SBO problem.
This has important consequences that set EML and SBO apart from the other problems.  Most importantly, the fact that $y^*$ is unavailable means that \emph{neither the ground-truth nor the model-side regret can be evaluated}.  This prevents EML and SBO from using standard DFL and SOP training objectives.  The only thing that \emph{can} be evaluated is the absolute cost of a candidate $\hat{y}$, namely $f(\hat{y})$.  Doing so, however, involves evaluating $h^*$, which is computationally challenging.

\paragraph{SOP/DFL vs SBO/EML}

The above analysis hints at a fundamental difference between EML/SBO on the one hand, and SOP/DFL on the other.
In EML and SBO, the only possible strategy is to (smartly) search or sample the landscape of candidates in search for high-quality solutions.
In contrast, in SOP and DFL, this is not necessary because it is known where some high-quality solutions $y_i^*$ are located.
In fact, \emph{in SOP pure exploration is impossible}, because $\theta^*$ is not available nor, therefore, 
the ground-truth cost $f(x_i, \hat{y}; \theta^*)$.

A potential point of similarity is related to the use of context information, i.e., the observable variables $x$. While context is a major focus in SOP and DFL, it is typically neglected in SBO/EML on the ground that a single problem instance is targeted. Taking advantage of context in these techniques would be not only possible, but beneficial: for example, by assuming that the complex component behaviour depends on observables, i.e. to have $h^*(x, y)$ rather than $h^*(y)$, it would be possible to learn surrogate models that are valid for multiple problem instances. This observation also draws a connection with contextual bandits approaches in SBO~\cite{krause2011contextual}. 
Note that, while re-using the learned model is an established practice in EML, currently this is done with no regard for the distribution of the observables $x$.

\paragraph{SBO/EML vs BBO}
EML is much closer to black-box optimisation (BBO) than to SOP/DFL, and SBO is even considered a BBO technique, albeit one that requires less function evaluations than other techniques.
What sets apart SBO and EML from classical BBO (e.g., Nelder-Mead) is their access to the structure of the surrogate model, which can be used to compute gradients, lower/upper bounds, and generically to perform constraint propagation.  In fact, reasoning on the surrogate model is the main focus in EML, which partly explains why the technique is not commonly associated to BBO. 
However, some SBO variants are also not considered a BBO technique, such as when gradient information of the true objective is included~\cite{Han2013ImprovingVS,SMT2019}.

\subsection{Open Problems and Research Directions}

The links between problems 
just identified open up several possibilities for knowledge transfer and future research.

\paragraph{Cross-fertilisation of SOP and DFL}
Several techniques used in SOP can be useful for DFL and vice versa.
For instance, the idea of imposing a minimum margin $\Delta$ on the regret achieved by incorrect outputs (Eq.~\ref{eq:sop-moia-with-margin}) transfers immediately to DFL, and may lead to improvements in generalisation.
Vice versa, bi-level formulations of the learning problem and implicit differentiation techniques commonly used in DFL can and are being adapted to SOP (compare \cite{niepert21imle}).

\paragraph{Cross-fertilisation of SBO and EML}
Knowledge transfer between SBO and EML also becomes evident.  One promising direction is to improve current EML algorithms, which assume all data to be provided upfront, by adapting iterative acquisition strategies from SBO, so to improve data quality, annotation costs, and sample complexity of the learning problem. As an extra challenge, such a strategy would need to deal with impact of the complex component on the problem constraints; however, on the plus side, any result in this area would be of immediate use for Bayesian Optimisation with constraints.
The focus on regret provides also a fresh perspective on a SBO (and potentially EML) training process (see Eq.~\eqref{eq:reg_sbo}): while classical approaches emphasise obtaining surrogates with high accuracy, achieving low regret is in fact equivalent and may enable the use of simpler (and cheaper) surrogate models.

\paragraph{Simple vs Accurate ML Models}
The focus on regret opens up the opportunity to use simpler models in all the considered techniques. The main idea is that it may be feasible to reach low regret by using low-variance models, as observed in \cite{elmachtoub2021smart}: for example, a linear regressor may lead to comparable performance to a moderately deep network. This is interesting since inference and/or training for such models is often much cheaper from a computational point of view, providing a mechanism to improve scalability. To the best of the authors knowledge, this line of research still needs to be properly investigated.

\paragraph{Scalability}
An open issue shared by all considered approaches is that of scalability, at training and/or inference time, which stems from the hybrid ML/combinatorial nature of the methods. At training time, it is necessary to solve an optimization problem for each dataset example (DFL, SOP) or to collect each new example (SBO): such problems can be NP-hard in some applications, limiting the dataset size that can be practically managed. At inference time, we need to reason on the ML model structure (SBO, EML): for many ML models (e.g., decision tree ensembles, neural networks, Gaussian processes) this is also an NP-hard problem, and makes dealing with large ML models a real challenge.

While scalability is still an open issue, due to its importance the area has been target of intense research. Strategies that have been investigated include: dimensionality reduction~\cite{Wang2016BayesianOI}, using sparse ML models~\cite{Yang2021SparseSG}, relying on relaxations \cite{mandi2020interior}, caching schemes \cite{mulamba2020discrete}, decompositions (e.g., \cite{DBLP:journals/corr/abs-1903-00958}). All solutions developed so far have been designed for individual techniques, and there is significant potential to improve scalability by simple cross-fertilisation of these approaches.

\paragraph{Prior Knowledge}
Prior knowledge can be employed to accelerate convergence at training time or to speed-up inference (or the solution process). This is in fact the main goal of all the hybrid approaches we presented, where prior knowledge is represented by the constraints and cost functions. That said, this research line is far from being exhausted: using actual Bayesian priors may be an option especially in SBO~\cite{Souza2021BayesianOW}; introducing redundant constraints may prevent the ML model from making critical mistakes \cite{borghesi2020combining}; known problem constraints may also be `pushed' in the ML model, for example in an attempt to prune it structure and make reasoning more computationally efficient.

\section{Related Topics}
\label{sec:related}

\paragraph{Generative models.}  Structured-output predictors can be viewed as a discriminative counterpart~ 
of probabilistic graphical models (PGMs).
Indeed, MAP queries in PGMs is equivalent to inference in structured-output prediction.  Work on structure learning of PGMs~\cite{drton2017structure} relies on strategies like integer programming~\cite{bartlett2017integer} and casting learning as a combinatorial optimisation problem~\cite{berg2014learning}.

\paragraph{Inverse optimisation.}  
Here the goal is to \emph{adapt} an initial 
combinatorial optimisation problem,
most often a linear program, so that it fits some example optimal solutions~\cite{ahuja2001inverse}.   
Historic approaches are only are concerned with adjusting the cost function itself but assume that the feasible region is fixed.  More recently, some approaches effectively perform constraint learning, as they do not require an initial COP and they can learn the feasible region too~\cite{barmann2017emulating,dong2018generalized}.

\paragraph{Learning to solve.}
In view of the widespread use of MIP, ML to accelerate MIP solving is a notable opportunity \cite{DBLP:journals/eor/BengioLP21}. 
Supervised and reinforcement learning is used to learn MIP branching rules \cite{DBLP:conf/aaai/ZarpellonJ0B21}, heuristic activation \cite{DBLP:journals/corr/abs-2103-10294}, column generation \cite{DBLP:journals/transci/MorabitD021}, and solution backdoors \cite{DBLP:journals/corr/abs-2110-08423}.

\paragraph{Constraint learning.} Automated constraint acquisition began to reach maturity with~\cite{de2018learning} and have recently
been extended to learning optimisation problems, cf.~\cite{kumar2020learning}.

\paragraph{Implicit structure learning.}
ML approaches for learning-to-reason~\cite{wang2019satnet} and learning-to-solve/optimise~\cite{amos2017optnet,ferber2020mipaal} subsume acquiring CO 
models.  These techniques aim to solve a 
COP
in relaxed form using (or within) neural architectures~\cite{ren2018adversarial,hu2018deep}.  We note that `within' neural architectures brings us back to DFL.

\section{Conclusion}
\label{sec:conc}

When COPs 
are not fully specified, it is natural to seek to leverage data.  In this regard, ML is a natural tool.
By adopting the common framework of regret minimisation, this paper not only surveyed ML for four important types of partially-specified optimisation problems -- structured-output prediction, decision-focused learning, empirical model learning and surrogate-based optimisation -- but brought them together in compatible formulations.  The commonalities and differences point to opportunities for cross-fertilisation.

\paragraph{Acknowledgements}
This research was partially supported by TAILOR, a project funded by EU Horizon 2020 research and innovation programme under grant number 952215.

\bibliographystyle{apalike}
\bibliography{references-nopages}
\end{document}